\documentclass{article} %
\usepackage{iclr2027_conference,times}

\usepackage{amsmath,amsfonts,bm}

\def\eqref#1{equation~\ref{#1}}

\def\1{\bm{1}}

\DeclareMathAlphabet{\mathsfit}{\encodingdefault}{\sfdefault}{m}{sl}
\SetMathAlphabet{\mathsfit}{bold}{\encodingdefault}{\sfdefault}{bx}{n}

\usepackage{hyperref}
\usepackage{url}
\usepackage{graphicx}
\usepackage{booktabs}

\DeclareUnicodeCharacter{03C0}{\ensuremath{\pi}}
\DeclareUnicodeCharacter{2080}{\ensuremath{_0}}

\title{If at First You Don't Succeed, Backtrack: Lightweight Recovery for Frozen VLAs}

\author{%
Thomas Landeg$^{1,2}$ \quad Thierry Tambe$^{1}$ \\
$^{1}$Stanford University \\
$^{2}$ETH Zurich \\
\texttt{\{tlandeg,ttambe\}@stanford.edu}
}

\iclrfinalcopy %
\begin{document}

\maketitle
\lhead{Preprint}

\begin{abstract}
Vision-Language-Action (VLA) models perform well on robotic manipulation, yet remain brittle once
their own actions carry them into states their training data barely covers. Recovering from such
failures usually requires recovery demonstrations, policy retraining, or a separate model at
inference time. We introduce \textbf{Backtrack}, a lightweight failure detection and recovery mechanism for
frozen VLAs that needs none of these. A small failure prediction head, trained only on
outcome-labeled rollouts of the frozen policy, reads the VLA's existing internal representations to
estimate whether the current episode will fail. When the predicted failure probability exceeds a
threshold, a controller pauses the policy, retraces the arm's recently achieved motion, and hands
control back to the unchanged policy so it can retry from an earlier configuration. The head adds
only 0.48--0.88\% to the policy's parameter count. We evaluate Backtrack on LIBERO with three VLAs
spanning two action-generation families: SmolVLA and $\pi$0.5, which use flow-matching action
experts, and the autoregressive $\pi$0-FAST. On LIBERO-Long, the suite of long-horizon multi-step
tasks and the one on which all three policies fail most often, Backtrack reduces each policy's
failure rate by 19--48\%, raising success by up to 7.4 percentage points; on the three shorter suites,
where control success is already 85--99\%, its effect is smaller. On MetaWorld, all twelve controller configurations
tested with SmolVLA improve on the unassisted policy.
\end{abstract}

\section{Introduction}

Recent advances in Vision-Language Models (VLMs) have spurred the development of
Vision-Language-Action (VLA) models \citep{black_05_2025, pertsch_fast_2025, shukor_smolvla_2025}.
These models typically use pretrained VLMs as a backbone, leveraging their visual and language representations to condition the generation of robot actions.
Yet even large VLAs trained on diverse, multi-embodiment robotics datasets remain brittle outside their training distribution: modest changes to camera viewpoint, robot initialization, or scene composition can sharply reduce success rates \citep{fei_libero-plus_2026}.
This challenge is particularly relevant for compact VLAs such as SmolVLA, whose VLM backbone benefits from broad vision-language pretraining, but whose robotics training consists of fewer than 30,000 episodes of community-contributed demonstrations collected primarily on a single robot embodiment \citep{shukor_smolvla_2025}.

Out-of-distribution states are not only something a VLA encounters; they are also something it can create. Robot imitation datasets are built primarily from successful demonstrations and contain comparatively little data showing how to recover once something has gone wrong \citep{dai_racer_2025, shukor_smolvla_2025}. For instance, when a VLA drops an object, knocks something over, or otherwise makes a mistake, it can therefore place itself in a state poorly represented in its training data.
Simply allowing more execution time leaves most failures unresolved: on LIBERO-Long,
doubling the episode step budget increases success by only 2.3--3.7 percentage points across
the three evaluated VLAs (Appendix~\ref{app:budget}).

One remedy is to give the policy the data it is missing: collect demonstrations of failures and their recoveries, then train on them \citep{dai_racer_2025}. 
But collecting recovery data is expensive, and the cost may need to be paid again for every new embodiment and task. Instead, we ask whether part of the distribution shift can be reversed directly. After a mistake, both the scene and the robot may have changed: objects may have been displaced, while the arm has moved into an unfamiliar configuration. Changes to the scene are difficult to undo, since restoring a displaced object is itself a manipulation problem. The arm configuration, however, is under our control. This suggests a simple recovery strategy: retrace the arm's motion along the path that led to failure, in the hope of returning toward an earlier configuration from which the unchanged policy can successfully retry.

\begin{figure}[t]
    \centering
    \includegraphics[width=\textwidth]{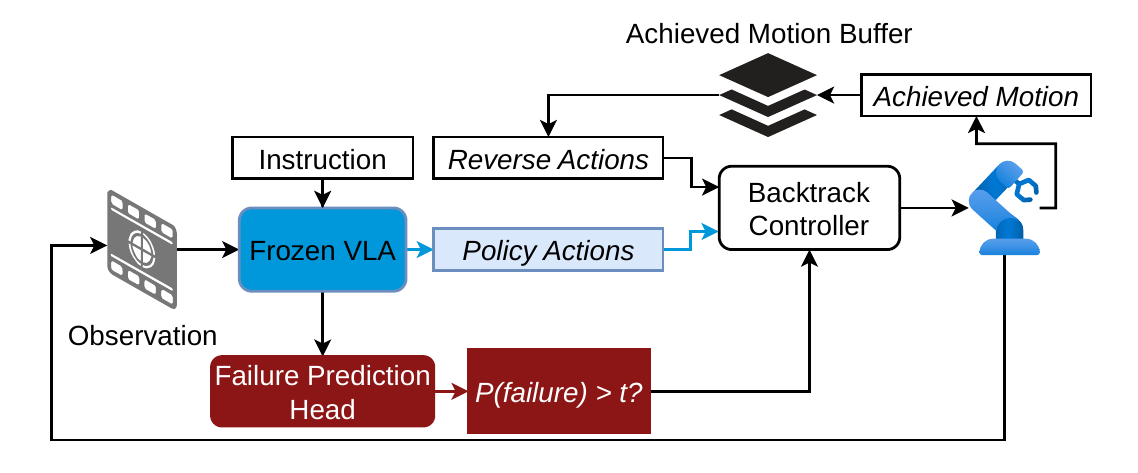}
    \caption{\textbf{Overview of Backtrack.}
    A small failure prediction head reads features already computed by the frozen VLA and estimates
    the probability that the current episode will fail. When the predicted failure probability
    exceeds the threshold $t$, the Backtrack controller suppresses the policy's proposed actions
    and instead retraces recent achieved arm motion stored in a buffer. Otherwise, the policy's
    actions are executed normally. The resulting state is observed by the unchanged VLA and the
    control loop continues. Achieved motion is added to the buffer during normal policy execution;
    motion executed during backtracking is not. The diagram abstracts over differences in action
    generation between the evaluated VLAs.}
    \label{fig:backtrack-overview}
\end{figure}

In this work, we introduce Backtrack, a lightweight failure detection and recovery mechanism for frozen VLAs (Figure~\ref{fig:backtrack-overview}). Similar to SAFE \citep{gu_safe_2025}, a small failure prediction head reads features the policy has already computed and estimates the probability that the current episode will end in failure; when that probability exceeds a threshold, a controller retraces the arm's recent motion and returns control to the policy. The only supervision Backtrack needs is whether autonomous rollouts of the frozen policy succeed or fail: no teleoperated recovery demonstrations, no policy retraining, and no second model at inference. The head adds just 0.48--0.88\% to the policy's parameters.

We evaluate Backtrack on three policies with two different action-generation mechanisms: SmolVLA \citep{shukor_smolvla_2025} and $\pi$0.5 \citep{black_05_2025} use flow-matching action experts, while $\pi$0-FAST \citep{pertsch_fast_2025} generates discrete action tokens autoregressively.
On LIBERO-Long \citep{liu_libero_2023}, the suite on which all three policies fail most often,
Backtrack reduces failure rates by 19\% to 48\%; averaged over all four LIBERO suites, the
reduction is 20--26\%.
On MetaWorld \citep{yu_meta-world_2020}, where we evaluate SmolVLA on a second benchmark
with a different embodiment, all twelve controller configurations we test improve on the
unassisted policy.

\section{Related Work}

\subsection{Vision-Language-Action Models}

VLAs adapt a pretrained vision-language backbone to output robot actions, inheriting its scene and
instruction understanding \citep{brohan_rt-2_2023, kim_openvla_2024}. They differ in how actions
are produced. One family attaches a separate action expert trained by flow matching or diffusion
and cross-attending to the backbone, as in SmolVLA \citep{shukor_smolvla_2025} and $\pi$0.5
\citep{black_05_2025}. The other discretizes actions into tokens that the backbone emits
autoregressively, as in $\pi$0-FAST \citep{pertsch_fast_2025}. A parallel line of work develops
compact VLAs and backbones for more efficient deployment
\citep{shukor_smolvla_2025, wang_vla-adapter_2025, marafioti_smolvlm_2025}. Backtrack attaches to
a frozen checkpoint and is agnostic to the action-generation architecture; we evaluate it on both
families.

\subsection{Failure Detection for VLAs}

Several lines of work detect signs of failure during policy execution. One uses a separate vision-language
model as a judge, either to classify success \citep{du_vision-language_2023} or to reason about
what went wrong \citep{duan_aha_2025, liu_reflect_2023}. Another uses signals derived from the policy itself: \citet{agia_unpacking_2025} monitor
consistency and progress at runtime, while SAFE \citep{gu_safe_2025} trains a lightweight
classifier on the policy's internal features and shows transfer across tasks and policies, including $\pi$0-FAST, but stops at detection. Backtrack builds
on this latter approach: we use the prediction to trigger a lightweight recovery
controller, measure what that intervention is worth across three policies, and test whether the
effect transfers to a second benchmark.

\subsection{Failure Recovery for VLAs}

Detection alone does not change the outcome of an episode; something must be done once a failure is
predicted. Retrying after a detected failure is not new. Bellman-Guided Retrials
\citep{du_err_2024} monitors progress with a value function trained on demonstrations and resamples
a strategy from the base policy when progress stalls, biasing away from behaviors that have
recently failed. Other work prompts a language model to reason its way to a corrective plan
\citep{lin_failsafe_2025, liu_reflect_2023}. 
These approaches introduce additional learned components or demonstrations to decide how to recover, but they do not provide a task-agnostic mechanism for returning the robot to a prior configuration while leaving the base policy unchanged.
Backtrack instead uses a task-agnostic recovery
primitive that requires no recovery demonstrations or learned recovery policy: the controller
retraces the arm's recent motion and then returns control to the unchanged VLA.

\section{Method}
\label{sec:method}

\subsection{Failure Prediction Head}
\label{sec:head}

We attach a \textit{Failure Prediction Head} (FPH) to the frozen policy. The FPH is a multi-head attention pooling (MAP) layer over the policy's own token features, followed by a three-layer feed-forward network (FFN) that outputs a single scalar: the predicted probability that the current episode will end in failure.
\paragraph{Where the features come from.}
The FPH consumes activations the policy already computes during normal action generation, so it
requires no additional forward pass through the backbone. We consider three feature taps: the prefix embeddings the VLM takes as \emph{input} (\texttt{pre}), the VLM's \emph{output} representation of that same prefix (\texttt{post}), and \texttt{post} concatenated with the action expert's own hidden states (\texttt{post\_expert}). Unless stated otherwise, all reported results use \texttt{post}.

$\pi$0-FAST has no separate action expert: the VLM emits the action tokens itself. Its \texttt{pre}
tap is the same as for the other two policies, the prefix embeddings the VLM takes as input. Its
\texttt{post} tap is the VLM's representation of that prefix after the prefill, just before the
first action token is generated.

\begin{table}[t]
\caption{Parameter count of the failure prediction head, relative to the frozen policy it
attaches to. All three heads read the \texttt{post} feature tap.}\label{tab:head-cost}
\begin{center}
\begin{tabular}{lrrr}
\toprule
\multicolumn{1}{c}{\bf POLICY} &\multicolumn{1}{c}{\bf PARAMS}
&\multicolumn{1}{c}{\bf HEAD PARAMS} &\multicolumn{1}{c}{\bf HEAD / POLICY} \\
\midrule
SmolVLA        & 450.0M  & 3.97M  & 0.88\% \\
$\pi$0-FAST    & 3450.0M & 17.35M & 0.50\% \\
$\pi$0.5       & 3616.8M & 17.35M & 0.48\% \\
\bottomrule
\end{tabular}
\end{center}
\end{table}

\paragraph{Parameter overhead.}
The head is evaluated once per action chunk and reuses activations already present in the forward pass, so its only marginal cost is the MAP layer and the FFN. Its size depends on the width of the features it reads rather than on the size of the policy, so the relative overhead is largest on the smallest policy: it adds at most \textbf{0.88\%} of the policy's parameters, on the 450M SmolVLA, and as little as 0.48\% on the 3.6B $\pi$0.5 (Table~\ref{tab:head-cost}).

\paragraph{Training.}
The head is trained on rollouts of the frozen policy, with each frame labeled by the outcome of the
episode it belongs to. Rollouts are collected across all four LIBERO suites, and we sample with
equal weight over the eight (suite, label) cells rather than from the natural distribution. This
matters because failures are distributed very unevenly across suites: for $\pi$0-FAST,
LIBERO-Long supplies 56\% of all failure frames and LIBERO-Object only 7\%, and
the imbalance is larger still for SmolVLA. Sampling from the raw distribution would give one head
per policy that is effectively specialized to the suite its policy fails on most, whereas we want a single head that serves all four. 
The amount of rollout data available differs across policies; Table~\ref{tab:headdata}
summarizes the rollout pools used for each head.

\begin{table}[t]
\caption{Rollout data behind each failure prediction head. The Episodes column gives the number of rollouts of the frozen policy pooled across all four
LIBERO suites; the failure rate is the fraction of those episodes that ended in failure. Sampled frames is the number drawn from that pool to train and validate the
head.}
\label{tab:headdata}
\begin{center}
\begin{tabular}{lrrr}
\toprule
\multicolumn{1}{c}{\bf POLICY} &\multicolumn{1}{c}{\bf EPISODES}
&\multicolumn{1}{c}{\bf FAILED} &\multicolumn{1}{c}{\bf SAMPLED FRAMES} \\
\midrule
SmolVLA     & 20{,}000 & 3{,}464 (17.3\%) & 100{,}000 \\
$\pi$0-FAST &  7{,}953 &   812 (10.2\%)   & 100{,}000 \\
$\pi$0.5    & 14{,}000 &   676 (4.8\%)    & 100{,}000 \\
\bottomrule
\end{tabular}
\end{center}
\end{table}

We train each head for 20{,}000 steps with Adam, batch size 32, learning rate $10^{-4}$ and weight
decay $10^{-5}$, with no additional positive-class weighting on top of the balanced sampler.

\subsection{Backtracking}
\paragraph{Trigger.}
Once the predicted failure probability exceeds a threshold, the policy is paused and the controller replays previously achieved arm motion in reverse, recomputing the failure probability at the same interval at which action chunks would otherwise be generated. In the fixed variant, the controller retreats for a predetermined number of steps; in the adaptive variant, it continues until the failure probability falls below a second, lower threshold, subject to minimum and maximum retreat lengths.
The intuition is that retracing the arm may undo part of the deviation that led to failure,
returning it toward a state better represented in the policy's training data.
Backtracking reverses the arm's motion without explicitly restoring the scene; displaced objects
may remain displaced, and contact during reversal may move them further.

\paragraph{Parameters.}
The controller has seven parameters. The \textit{threshold} $t$ is the predicted failure probability that triggers a retreat. A \textit{fixed} retreat reverses a set number of steps $s$; an \textit{adaptive} retreat continues until the probability falls below the \textit{restart threshold} $a$, bounded by $s_{\min}$ and $s_{\max}$ steps. The \textit{cooldown} $\mathit{cd}$ is the number of steps after a retreat during which the trigger is suppressed, so the controller does not immediately re-fire on the state it has just left. The \textit{retreat cap} $K$ limits how many retreats may be initiated per episode.
The configurations selected for the main experiments (Appendix~\ref{app:sweeps}) are: SmolVLA, adaptive with $t{=}0.97$, $a{=}0.2$, 50--150 steps; $\pi$0.5, fixed with $t{=}0.99$, $s{=}50$; and $\pi$0-FAST, adaptive with $t{=}0.90$, $a{=}0.3$, 25--100 steps. All three use $\mathit{cd}{=}50$ and $K{=}3$.

\paragraph{What is replayed.}
During normal operation we append every executed motion to a buffer. These are not the actions commanded by the policy. The policy emits deltas relative to the arm's current pose, but if the arm cannot move as commanded, for instance because the motion is blocked by an object or by the work surface, recording the command would cause the reversal to undo a motion that never happened. We therefore record the \emph{achieved} delta, that is, the difference between the arm's current and previous pose. The gripper is replayed with the arm: on LIBERO, where gripper commands are absolute, the recorded commands are re-issued in reverse order, so retreating past a grasp re-opens the gripper; on MetaWorld, where they are incremental, they are negated.

\paragraph{The return is approximate.}
Reversing recorded deltas does not restore the arm exactly. Although we record achieved rather
than commanded motion, the reversed deltas are themselves executed approximately because of
the embodiment's tracking dynamics. The arm therefore returns to a neighborhood of the earlier
state rather than to the state itself.

\section{Experimental Setup}
\label{sec:setup}

\subsection{Benchmark and protocol}
We evaluate primarily on LIBERO \citep{liu_libero_2023}, using its four ten-task suites:
LIBERO-Spatial, LIBERO-Object, LIBERO-Goal, and LIBERO-Long.
To check that the method is not specific to one benchmark we additionally
evaluate on MetaWorld \citep{yu_meta-world_2020}, a separate manipulation benchmark with a
different simulator, task set, and embodiment.
The MetaWorld experiments use SmolVLA with the \texttt{pre} feature tap.

We build on the LeRobot library \citep{cadene_lerobot_2024}, which provides the policy
implementations, the LIBERO environment wrappers, and the evaluation harness. We implement the
failure prediction head, the backtracking controller, and our evaluation extensions in a fork of
LeRobot.

\paragraph{Protocol.}
Our main LIBERO evaluation uses five seeds and 50 episodes per task. We use one controller
configuration per policy, selected on LIBERO-Long and transferred unchanged to the other
three suites; the grid and selection procedure are given in Appendix~\ref{app:sweeps}. Control and
treatment use the same step budget, and retreat steps are charged against it. We report uplift,
the treatment minus control success rate in percentage points, paired by seed, with 95\%
confidence intervals over the five per-seed differences.

\paragraph{Step budget.}
We evaluate at twice the per-suite step limit set by LeRobot's LIBERO wrappers, giving
560 steps for LIBERO-Spatial and LIBERO-Object, 600 for LIBERO-Goal,
and 1040 for LIBERO-Long. The default limits are set just above the longest training
demonstration in each suite, leaving as few as fifteen steps of slack on LIBERO-Long.
This leaves little room for a policy to deviate and subsequently recover before reaching the episode limit.
We double these limits because a retreat is charged against the same budget, and we apply the same budget to control
and treatment throughout. The original benchmark does not specify per-suite episode limits, using a
single 600-step cap for its own training-time evaluation \citep{liu_libero_2023}.

\subsection{Policies}

\paragraph{SmolVLA.}
SmolVLA \citep{shukor_smolvla_2025} is a compact VLA that truncates SmolVLM2 \citep{marafioti_smolvlm_2025} and adds a flow-matching action expert, which cross-attends to the VLM's KV cache. In their LIBERO evaluation, \citet{shukor_smolvla_2025} use 10 flow-matching steps and regenerate an action chunk after every executed action. We change two inference settings. First, we use only 2 flow-matching steps, since our ablation in Appendix~\ref{app:steps} shows this only slightly degrades performance on LIBERO and improves it on MetaWorld \citep{yu_meta-world_2020}. Second, we execute 10 actions from each generated action chunk before regenerating, which substantially reduces the number of policy forward passes required during evaluation.

\paragraph{$\pi$0.5.}
$\pi$0.5 is the largest policy we evaluate, with 3.6B parameters \citep{black_05_2025}. Like SmolVLA, it uses a VLM backbone with a flow-matching action expert, but at roughly eight times the parameter count.

\paragraph{$\pi$0-FAST.}
$\pi$0-FAST is the $\pi$0 VLA \citep{black__2025} with its flow-matching action expert replaced
by autoregressive decoding of discrete action tokens \citep{pertsch_fast_2025}. Unlike the other
two policies we evaluate, it has no separate action expert: the VLM emits the action tokens itself.

\section{Results}
\label{sec:results}

\subsection{Recovery}

Table~\ref{tab:main} reports the main result. On LIBERO-Long, Backtrack reduces the failure
rates of SmolVLA, $\pi$0.5 and $\pi$0-FAST by 19.0\%, 30.3\% and 48.2\%, respectively. Every 95\% confidence interval excludes zero.
Aggregated over all four LIBERO suites, the three policies sit much closer
together, at 20.3\%, 26.2\% and 19.6\%, again with every interval excluding zero.

The spread across policies is much wider on LIBERO-Long, at 19--48\%, than over the
benchmark as a whole, at 19.6--26.2\%, and the ordering is not preserved between the two:
$\pi$0-FAST has the largest reduction on the long-horizon suite and the smallest reduction
overall. Recovery rate thus depends on the evaluation suite; these three policies do not establish a
relationship between recovery rate and base success rate.

\begin{table}[t]
\caption{Recovery on the long-horizon LIBERO-Long suite, and aggregated over all four
LIBERO suites. Five seeds (1000--5000), 50 episodes per task. Uplift is in percentage points and
intervals are 95\% confidence intervals over the five per-seed differences. Recovery rate is the
uplift divided by the base failure rate, measuring the net relative reduction in failures.}
\label{tab:main}
\begin{center}
\begin{tabular}{llrrrr}
\toprule
\multicolumn{1}{c}{\bf SUITE} &\multicolumn{1}{c}{\bf POLICY} &\multicolumn{1}{c}{\bf CONTROL}
&\multicolumn{1}{c}{\bf UPLIFT} &\multicolumn{1}{c}{\bf 95\% CI} &\multicolumn{1}{c}{\bf RECOVERY} \\
\midrule
LIBERO-Long & SmolVLA     & 61.00 & $+7.40$ & $[+4.52, +10.28]$ & 19.0\% \\
                    & $\pi$0.5    & 94.72 & $+1.60$ & $[+0.06, +3.14]$  & 30.3\% \\
                    & $\pi$0-FAST & 85.32 & $+7.08$ & $[+5.91, +8.25]$  & 48.2\% \\[4pt]
all four suites     & SmolVLA     & 83.26 & $+3.40$ & $[+2.04, +4.76]$  & 20.3\% \\
                    & $\pi$0.5    & 96.98 & $+0.79$ & $[+0.05, +1.53]$  & 26.2\% \\
                    & $\pi$0-FAST & 91.31 & $+1.70$ & $[+0.97, +2.43]$  & 19.6\% \\
\bottomrule
\end{tabular}
\end{center}
\end{table}

\paragraph{Selection.}
Each policy's configuration was selected on LIBERO-Long using seeds 1000--3000, where
Backtrack has the most room to improve performance, and transferred unchanged to the other three
suites. Seeds 4000--5000 played no part in selection; on them, the selected configurations give
$+7.20$\,pp for SmolVLA (vs.\ $+7.40$ pooled), $+0.70$\,pp for $\pi$0.5 (vs.\ $+1.60$) and
$+7.40$\,pp for $\pi$0-FAST (vs.\ $+7.08$). For $\pi$0.5 and $\pi$0-FAST, selection was run on a
different feature tap from the one reported (\texttt{post\_expert} and \texttt{pre},
respectively), and the configurations were carried over to \texttt{post} unchanged; subsequent
\texttt{post}-head comparisons found no better configuration among those tested.
Appendix~\ref{app:sweeps} gives the complete selection procedures, configuration grids, and
post-selection checks.

\subsection{Where recovery pays}

Table~\ref{tab:persuite} breaks the aggregate down by suite. The three short suites sit closer to
ceiling, between 84.6\% and 99.2\% control success, leaving less room for improvement. The
long-horizon suite provides the largest uplift for all three policies.

A simple accounting identity illustrates the tradeoff. Let $S$ and $F$ be the baseline success
and failure proportions. Recovering a fraction $c$ of failures while disrupting a fraction $b$ of
successes changes the success rate by $cF-bS$, or $100(cF-bS)$ percentage points. For $c>0$,
this is positive when $b/c<F/S$. On $\pi$0.5's LIBERO-Object, $S/F$ is 124:
disrupting just 1\% of baseline successes would outweigh recovering every baseline failure.
On LIBERO-Long for SmolVLA, $S/F$ is only 1.6, allowing a larger disruption rate
for the same fraction of failures recovered.

This motivates selecting configurations on LIBERO-Long, where baseline failure rates
are highest. However, $b$ and $c$ may also vary across suites, and the configurations are transferred
unchanged. These results therefore do not isolate the effects of baseline failure rate, task
characteristics, and configuration selection.

\begin{table}[t]
\caption{Per-suite control success rate and uplift, five seeds, 50 episodes per task.}
\label{tab:persuite}
\begin{center}
\begin{tabular}{lrrrrrrrr}
\toprule
&\multicolumn{2}{c}{\bf SPATIAL} &\multicolumn{2}{c}{\bf OBJECT} &\multicolumn{2}{c}{\bf GOAL}
&\multicolumn{2}{c}{\bf LONG} \\
\cmidrule(lr){2-3}\cmidrule(lr){4-5}\cmidrule(lr){6-7}\cmidrule(lr){8-9}
\multicolumn{1}{c}{\bf POLICY} & ctrl & upl & ctrl & upl & ctrl & upl & ctrl & upl \\
\midrule
SmolVLA     & 84.64 & $+5.12$ & 94.72 & $-0.20$ & 92.68 & $+1.28$ & 61.00 & $+7.40$ \\
$\pi$0.5    & 98.24 & $+0.16$ & 99.20 & $+0.16$ & 95.76 & $+1.24$ & 94.72 & $+1.60$ \\
$\pi$0-FAST & 90.92 & $-0.36$ & 96.80 & $-0.20$ & 92.20 & $+0.28$ & 85.32 & $+7.08$ \\
\bottomrule
\end{tabular}
\end{center}
\end{table}

\subsection{Detection}

Across 10{,}000 Backtrack episodes per policy, spanning all four suites and five seeds, success
conditioned on no trigger is 99.71\% for SmolVLA, 99.99\% for $\pi$0.5 and 99.94\% for
$\pi$0-FAST: 29 failures occur without a trigger in 30{,}000 episodes. These comprise 23 SmolVLA
failures on LIBERO-Long, one $\pi$0.5 failure on LIBERO-Goal, and five
$\pi$0-FAST failures on LIBERO-Object. Among triggered episodes, the mean first-trigger
time ranges from 11.0\% to 22.2\% of the budget across the twelve policy--suite pairs,
leaving about 810--870 steps on LIBERO-Long. These results show that few failures go unflagged; they do not by themselves establish how often
the intervention was necessary.

\paragraph{Offline discrimination.}
For $\pi$0.5, we also score the head offline on 2{,}772 held-out rollouts of the frozen policy
(135 failures), using a sample of frames from each rollout. It separates failing from successful
episodes with an episode-level AUROC of 0.977 (frame-level 0.908). At the deployed threshold
$t=0.99$, it flags 80\% of failing episodes at a 2.0\% false-alarm rate; lowering the threshold to
$t=0.9$ raises detection to 96\% but triples the false-alarm rate to 6.1\%. The
\texttt{pre} and \texttt{post\_expert} taps reach similar AUROCs (0.980 and 0.987), consistent
with the small closed-loop differences between taps (Section~\ref{sec:ablations}).

\subsection{A second benchmark}
\label{sec:metaworld}

The results above are all on LIBERO with the \texttt{post} feature tap. We also evaluate
Backtrack on MetaWorld with SmolVLA and the \texttt{pre} feature tap, using a head trained on
rollouts from MetaWorld's medium-difficulty tasks and evaluated on all 50 tasks, spanning every
difficulty level.

SmolVLA's control success rate on MetaWorld is 72.32\%. We swept twelve controller configurations,
five seeds each. \emph{Every one of them improves on control}, with observed improvements ranging
from $+2.44$ to $+5.20$ percentage points; eleven of the twelve seed-paired 95\% confidence
intervals exclude zero (Appendix Table~\ref{tab:metaworld-sweep}). The largest observed improvement
comes from a fixed 150-step retreat with a 100-step cooldown at $t=0.9$, reaching 77.52\%
success and a recovery rate of 18.8\%.

\subsection{Ablations}
\label{sec:ablations}

\paragraph{Which features to read.}
On both flow-matching policies, we find no resolvable benefit from adding the action expert's hidden states to the VLM representation (Appendix~\ref{app:feature-taps}). On $\pi$0.5, where we evaluate this comparison under the full evaluation protocol, \texttt{post} achieves 96.32\% versus 96.60\% for \texttt{post\_expert}, a paired difference of $-0.28$\,pp with a 95\% interval of $[-0.97, +0.41]$. Since \texttt{post} is available before the action chunk is generated and can therefore support parallel failure prediction and action generation, we use it for the main LIBERO evaluations. On $\pi$0-FAST, which has no separate action expert, the \texttt{post}-minus-\texttt{pre} difference is likewise unresolved ($+1.40$\,pp, 95\% interval $[-3.0, +5.8]$).

\paragraph{Threshold and retreat rule.} %
On $\pi$0.5, Table~\ref{tab:ablate} sweeps the trigger threshold under both fixed and adaptive retreat. The headline
configuration, a fixed 50-step retreat with a 50-step cooldown at $t=0.99$, has the highest
observed success rate at 96.32\%, followed by 96.24\% at $t=0.97$.

Fixed and adaptive retreat behave differently depending on the trigger threshold. At the lower
thresholds, fixed retreat is clearly better: adaptive minus fixed is $-2.76$\,pp at $t=0.90$ and
$-2.00$\,pp at $t=0.95$, with both intervals excluding zero. At $t=0.99$ and $t=0.995$, however,
the difference is no longer resolvable. Notably, adaptive retreat at $t=0.90$ is the only
configuration in the sweep that performs significantly worse than control, at $-1.92$\,pp with
$[-3.14, -0.70]$, while fixed retreat at the same threshold significantly improves over control,
at $+0.84$\,pp with $[+0.11, +1.57]$.

One possible explanation is that the two rules differ in how much of the episode budget they consume
when the detector fires frequently. At a low threshold, an adaptive retreat of up to 150 steps can
spend substantially more of the budget per trigger than a fixed 50-step retreat. Fewer triggers
at higher thresholds could reduce this cost. The observed narrowing of the gap is consistent
with this explanation, but does not isolate it from the effects of intervention timing or state.

\begin{table}[t]
\caption{Threshold and retreat rule on $\pi$0.5, LIBERO-Long, \texttt{post} head, five
seeds, 50 episodes per task, paired by seed. Control success is 94.72\%, a failure rate of
5.28\%. The first two blocks sweep the trigger threshold under each retreat rule; the third
varies retreat depth at $t{=}0.99$. The adaptive rule retreats between 50 and 150 steps, stopping
once the predicted failure probability falls below $0.2$.}
\label{tab:ablate}
\begin{center}
\begin{tabular}{lrrrr}
\toprule
\multicolumn{1}{c}{\bf CONFIGURATION} &\multicolumn{1}{c}{\bf SR} &\multicolumn{1}{c}{\bf UPLIFT}
&\multicolumn{1}{c}{\bf 95\% CI} &\multicolumn{1}{c}{\bf RECOVERY} \\
\midrule
\multicolumn{5}{l}{\emph{Threshold, fixed retreat} ($s{=}50$, $cd{=}50$)} \\
$t{=}0.90$            & 95.56 & $+0.84$ & $[+0.11, +1.57]$ & 15.9\% \\
$t{=}0.95$            & 95.00 & $+0.28$ & $[-1.24, +1.80]$ &  5.3\% \\
$t{=}0.97$            & 96.24 & $+1.52$ & $[+0.02, +3.02]$ & 28.8\% \\
$t{=}0.99$ (headline) & 96.32 & $+1.60$ & $[+0.06, +3.14]$ & 30.3\% \\
$t{=}0.995$           & 95.56 & $+0.84$ & $[-0.36, +2.04]$ & 15.9\% \\[4pt]
\multicolumn{5}{l}{\emph{Threshold, adaptive retreat} (50--150 steps, $cd{=}50$)} \\
$t{=}0.90$            & 92.80 & $-1.92$ & $[-3.14, -0.70]$ & $-36.4$\% \\
$t{=}0.95$            & 93.00 & $-1.72$ & $[-3.92, +0.48]$ & $-32.6$\% \\
$t{=}0.99$            & 95.60 & $+0.88$ & $[-0.81, +2.57]$ & 16.7\% \\
$t{=}0.995$           & 95.32 & $+0.60$ & $[-2.06, +3.26]$ & 11.4\% \\[4pt]
\multicolumn{5}{l}{\emph{Retreat depth, fixed} ($t{=}0.99$)} \\
$s{=}50$, $cd{=}50$ (headline)   & 96.32 & $+1.60$ & $[+0.06, +3.14]$ & 30.3\% \\
$s{=}100$, $cd{=}100$            & 95.44 & $+0.72$ & $[-0.59, +2.03]$ & 13.6\% \\
\bottomrule
\end{tabular}
\end{center}
\end{table}

\paragraph{Controller sensitivity across policies.}
Table~\ref{tab:ablate-smolvla} repeats the controller ablations on SmolVLA, where the failure rate
is substantially larger than on $\pi$0.5 (39.00\% versus 5.28\%). Every configuration tested
improves over control. The threshold sweep peaks at the reported $t=0.97$, but the response is
relatively broad: thresholds from $0.90$ to $0.99$ yield uplifts between $+4.60$ and $+7.40$\,pp.
Of the alternative thresholds, only $t=0.995$ resolvably underperforms the headline configuration,
while still yielding a $+4.00$\,pp uplift over control. 

On SmolVLA at $t=0.97$, adaptive retreat has a $2.24$\,pp higher mean success rate than
fixed 50-step retreat, but the 95\% interval $[-0.21,+4.69]$ includes zero. Fixed retreat
performs better on $\pi$0.5 at lower thresholds; at higher thresholds, its difference from
adaptive retreat is unresolved. No single retreat rule performs best across the tested policies and thresholds.

\begin{table}[t]
\caption{Threshold and retreat rule on SmolVLA, LIBERO-Long, \texttt{post} head, five
seeds, 50 episodes per task, paired by seed. Control success is 61.00\%, a failure rate of
39.00\%. The first block sweeps the trigger threshold under SmolVLA's reported adaptive rule,
which retreats between 50 and 150 steps, stopping once the predicted failure probability falls
below $0.2$; the second holds the threshold at the reported $t{=}0.97$ and substitutes a fixed
retreat. Fire rate is the proportion of episodes in which the controller triggers at least once.
Unlike for $\pi$0.5 (Table~\ref{tab:ablate}), we did not sweep the threshold under the fixed rule.}
\label{tab:ablate-smolvla}
\begin{center}
\begin{tabular}{lrrrrr}
\toprule
\multicolumn{1}{c}{\bf CONFIGURATION} &\multicolumn{1}{c}{\bf SR} &\multicolumn{1}{c}{\bf UPLIFT}
&\multicolumn{1}{c}{\bf 95\% CI} &\multicolumn{1}{c}{\bf RECOVERY}
&\multicolumn{1}{c}{\bf FIRE} \\
\midrule
\multicolumn{6}{l}{\emph{Threshold, adaptive retreat} (50--150 steps, $cd{=}50$)} \\
$t{=}0.90$            & 65.76 & $+4.76$ & $[+1.19, +8.33]$  & 12.2\% & 49.8\% \\
$t{=}0.95$            & 66.44 & $+5.44$ & $[+2.78, +8.10]$  & 13.9\% & 47.5\% \\
$t{=}0.97$ (headline) & 68.40 & $+7.40$ & $[+4.52, +10.28]$ & 19.0\% & 42.6\% \\
$t{=}0.99$            & 65.60 & $+4.60$ & $[+2.43, +6.77]$  & 11.8\% & 42.5\% \\
$t{=}0.995$           & 65.00 & $+4.00$ & $[+0.69, +7.31]$  & 10.3\% & 39.2\% \\[4pt]
\multicolumn{6}{l}{\emph{Retreat rule and depth, fixed} ($t{=}0.97$)} \\
$s{=}50$, $cd{=}50$   & 66.16 & $+5.16$ & $[+4.29, +6.03]$  & 13.2\% & 43.3\% \\
$s{=}100$, $cd{=}100$ & 67.04 & $+6.04$ & $[+3.90, +8.18]$  & 15.5\% & 44.0\% \\
\bottomrule
\end{tabular}
\end{center}
\end{table}

\paragraph{Number of retreats.}
We vary the retreat cap $K$ from one to five on $\pi$0.5, with three seeds per cap. The mean
success rates span only 0.60\,pp, and no cap is resolvably better than another
(Appendix~\ref{app:cap}). All other experiments use $K{=}3$.

\section{Discussion and Limitations}
\label{sec:discussion}

\subsection{What limits recovery?}

Retreating returns the arm toward an earlier configuration after a predicted failure. A retry may
differ because the arm returns only approximately, the scene has changed, or the policy samples
different actions. Our experiments do not isolate these sources of variation. The detector supplies
the trigger, while backtracking determines where the policy resumes; success depends on whether
the unchanged policy can complete the task from that resulting state.

Backtrack deliberately uses a simple recovery mechanism: it requires no recovery demonstrations
or retraining of the VLA, and does not learn a recovery policy or generate a new strategy. That this
intervention reduces failure rates by 19--48\% on LIBERO-Long suggests how much can be gained
from return-and-retry alone, while the remaining failures leave headroom for recovery mechanisms that
can actively change the policy's strategy after failure.

\subsection{Limitations}

\paragraph{Simulation only.}
Every result here is in simulation, and we do not validate on a physical robot. MetaWorld adds a second benchmark, but it is also simulated and was run only
with SmolVLA.

\paragraph{Reversibility.}
Backtrack restores neither object poses nor the full environment state, and executing the return
trajectory can itself change the scene. It cannot recover failures that have made the task
physically impossible, and it may not help when recovery requires a new manipulation strategy.

\section{Conclusion}

We introduced Backtrack, a lightweight failure detection and recovery mechanism for frozen VLAs.
Backtrack trains only a small failure prediction head on rollouts of the unchanged policy and
requires no recovery demonstrations or VLA retraining. At inference time, it uses the policy's own
representations to detect likely failure and retraces recent arm motion before returning control
to the policy. The head adds less than 1\% to the policy's parameter count, even on the
450M-parameter SmolVLA, and requires no separate model at inference.

On LIBERO-Long, the suite on which all three VLAs fail most often, Backtrack reduces failure
rates by 19--48\%; on the three shorter suites, where control success is already 85--99\%, its
effect is smaller.
The method also improves performance on a second benchmark: on MetaWorld, every controller
configuration tested with SmolVLA, the smallest policy we evaluate, improves on the unassisted
policy, as does every SmolVLA configuration tested on LIBERO-Long. These results suggest that a
meaningful fraction of VLA failures can be recovered with a simple return-and-retry mechanism,
without collecting recovery demonstrations or modifying the underlying policy, and that the
benefit is largest where the base policy fails most often.

\subsection*{AI use statement}

We used generative AI tools to assist with software development,
literature review, feedback on research methodology and experiments, discussion of
experimental results, and drafting and editing parts of the text. The authors made all
methodological decisions, verified all analyses and AI-generated code, and take full
responsibility for the content of this paper.

\subsection*{Reproducibility statement}

The paper specifies the failure prediction head and recovery controller, and
Section~\ref{sec:setup} describes the policies, evaluation protocol, and step budgets.
Appendix~\ref{app:sweeps} reports the hyperparameter sweeps and configuration-selection
procedure, with additional ablations and implementation details provided throughout the
appendix.

\subsubsection*{Acknowledgments}
This work used Marlowe~\citep{kapfer_marlowe_2025},
Stanford University's GPU-based Computational Instrument, supported by Stanford HAI and
Stanford Research Computing.

\bibliography{zotero}
\bibliographystyle{iclr2027_conference}

\appendix
\section{Appendix}

\subsection{Hyperparameter sweeps and configuration selection}
\label{app:sweeps}
Each policy is evaluated with a single controller configuration, selected on
LIBERO-Long and transferred unchanged to the other three suites. The paragraphs below
describe the sweep used to select each reported configuration; for $\pi$0-FAST and $\pi$0.5,
selection was performed using a different feature tap from the one used for final evaluation.

\paragraph{SmolVLA.} Three configurations, differing only in the trigger threshold
$t \in \{0.90, 0.97, 0.99\}$, with the retreat rule held fixed (adaptive, restart threshold $a = 0.2$, minimum 50
steps, maximum 150, cooldown 50). Three seeds at 10 episodes per task. The configuration with
$t = 0.97$ had the highest mean success rate in this sweep.
At that sample size the per-arm intervals span roughly $\pm 20$\,pp, so this comparison cannot
separate the three candidates and the choice should be read as weakly determined rather than
tuned. We therefore re-evaluate the threshold at the full protocol in
Section~\ref{sec:ablations}: five seeds and 50 episodes per task, over five settings spanning
$0.90$--$0.995$. There $t = 0.97$ remains the highest of the five, the four alternatives cluster
$2$--$3.4$\,pp below it, and only $t = 0.995$ is resolvably worse. The retreat parameters
($a = 0.2$, minimum 50, maximum 150, cooldown 50) were held fixed throughout and never tuned.

\paragraph{$\pi$0-FAST.} Six configurations on LIBERO-Long, five seeds at 10 episodes per
task: adaptive retreats at $t \in \{0.80, 0.85, 0.90\}$ with $a = 0.2$, minimum 50, maximum 150
and cooldown 50; an adaptive retreat at $t = 0.90$ with $a = 0.3$, minimum 25, maximum 100 and
cooldown 50; and fixed retreats of 50 steps (cooldown 50) and 100 steps (cooldown 100), both at
$t = 0.90$. The adaptive $t = 0.90$, $a = 0.3$ configuration had the highest mean success rate
on selection seeds 1000--3000; seeds 4000--5000 were held out.

This selection was run on the \texttt{pre} head, and the configuration was carried over unchanged
to the \texttt{post} head used for final evaluation. The two taps were compared separately
(Section~\ref{sec:ablations}), but that comparison does not establish that controller rankings are
preserved across representations; the \texttt{post}-head checks below address this directly.

After selection, we ran two threshold sweeps on the \texttt{post} head at the full protocol, five
seeds and 50 episodes per task, both paired by seed against the headline control
(Table~\ref{tab:pi0fast-post}). The first holds the reported retreat shape fixed (adaptive,
$a = 0.3$, minimum 25, maximum 100, cooldown 50) and varies the threshold over
$t \in \{0.95, 0.99, 0.995\}$, with the reported $t = 0.90$ arm as the fourth point. All three
alternatives improve over control, but each is resolvably worse than the reported configuration
when paired by seed: $-4.92$\,pp $[-6.48, -3.36]$ at $t = 0.95$, $-1.56$\,pp $[-3.00, -0.12]$ at
$t = 0.99$, and $-2.96$\,pp $[-4.82, -1.10]$ at $t = 0.995$. The second sweep uses the retreat
shape of the SmolVLA and $\pi$0.5 threshold sweeps (adaptive, $a = 0.2$, minimum 50, maximum 150,
cooldown 50) over $t \in \{0.90, 0.95, 0.99, 0.995\}$. Its best point, $+4.40$\,pp at $t = 0.99$,
is below the reported configuration. At the shared $t = 0.90$, the two shapes differ by
$6.32$\,pp. Threshold rankings also depend on retreat shape: $t = 0.90$ is the best threshold
under the reported shape and the worst under the other. The reported configuration is the best of the eight \texttt{post}-head configurations
evaluated, although only these two retreat shapes were tested on \texttt{post}.
These sweeps were run after the configuration had been selected and on the same seeds as the
headline arm, so we use them only as a check, not to select or re-rank configurations.

\begin{table}[t]
\caption{Threshold and retreat shape on $\pi$0-FAST, LIBERO-Long, \texttt{post} head, five
seeds, 50 episodes per task, paired by seed. Control success is 85.32\%, a failure rate of
14.68\%. The first block varies the threshold under the reported retreat shape; the second
varies it under the retreat shape used for the SmolVLA and $\pi$0.5 threshold sweeps
(Tables~\ref{tab:ablate} and~\ref{tab:ablate-smolvla}). $a$ is the restart threshold. Fire rate is
the proportion of episodes in which the controller triggers at least once.}
\label{tab:pi0fast-post}
\begin{center}
\begin{tabular}{lrrrrr}
\toprule
\multicolumn{1}{c}{\bf CONFIGURATION} &\multicolumn{1}{c}{\bf SR} &\multicolumn{1}{c}{\bf UPLIFT}
&\multicolumn{1}{c}{\bf 95\% CI} &\multicolumn{1}{c}{\bf RECOVERY}
&\multicolumn{1}{c}{\bf FIRE} \\
\midrule
\multicolumn{6}{l}{\emph{Threshold, reported shape} (adaptive, $a{=}0.3$, 25--100 steps, $cd{=}50$)} \\
$t{=}0.90$ (headline) & 92.40 & $+7.08$ & $[+5.91, +8.25]$ & 48.2\% & 33.7\% \\
$t{=}0.95$            & 87.48 & $+2.16$ & $[+0.45, +3.87]$ & 14.7\% & 31.4\% \\
$t{=}0.99$            & 90.84 & $+5.52$ & $[+3.42, +7.62]$ & 37.6\% & 20.8\% \\
$t{=}0.995$           & 89.44 & $+4.12$ & $[+2.38, +5.86]$ & 28.1\% & 19.6\% \\[4pt]
\multicolumn{6}{l}{\emph{Threshold, SmolVLA/$\pi$0.5 shape} (adaptive, $a{=}0.2$, 50--150 steps, $cd{=}50$)} \\
$t{=}0.90$            & 86.08 & $+0.76$ & $[-0.96, +2.48]$ &  5.2\% & 38.0\% \\
$t{=}0.95$            & 88.32 & $+3.00$ & $[+1.02, +4.98]$ & 20.4\% & 29.8\% \\
$t{=}0.99$            & 89.72 & $+4.40$ & $[+2.55, +6.25]$ & 30.0\% & 21.7\% \\
$t{=}0.995$           & 88.92 & $+3.60$ & $[+0.52, +6.68]$ & 24.5\% & 20.1\% \\
\bottomrule
\end{tabular}
\end{center}
\end{table}

\paragraph{$\pi$0.5.} Four configurations on LIBERO-Long, five seeds at 50 episodes per
task: fixed retreats of 50 steps (cooldown 50) and 100 steps (cooldown 100) at $t = 0.99$, a fixed
50-step retreat at $t = 0.995$, and an adaptive retreat at $t = 0.99$ spanning 25 to 100 steps with
a cooldown of 50 and a restart threshold of $0.3$. The fixed 50-step, 50-cooldown configuration
at $t = 0.99$ had the highest mean success rate on selection seeds 1000--3000; held-out seeds
4000--5000 give $+0.70$\,pp against a pooled $+1.60$\,pp. This grid varied the retreat rule and its depth and compared
the threshold only against $0.995$; the full threshold sweep is reported in
Section~\ref{sec:ablations}. The sweep read the \texttt{post\_expert} tap. All four configurations
were re-run on \texttt{post}, where the same configuration again wins on seeds 1000--3000
($+2.20$\,pp against $+1.07$ to $+1.33$\,pp).

\paragraph{Held-out reporting.} All three configurations were selected using seeds 1000--3000
only, so Section~\ref{sec:results} also reports their uplift on seeds 4000--5000, which played no
part in selection. All three continue to improve over control on these seeds. SmolVLA's uplift is
essentially unchanged, at $+7.20$\,pp against a pooled $+7.40$; $\pi$0-FAST's is $+7.40$\,pp
against a pooled $+7.08$; and $\pi$0.5's remains positive at $+0.70$\,pp against a pooled
$+1.60$, a reduction consistent with the narrow margins separating its candidate configurations.
These estimates evaluate configurations fixed using seeds 1000--3000; they do not quantify
variation in the selection procedure across different splits.

\subsection{Flow-matching steps}
\label{app:steps}

SmolVLA is evaluated by \citet{shukor_smolvla_2025} with ten flow-matching steps; we use two
throughout. On MetaWorld, five seeds at 500 episodes each, two steps is \emph{better} than ten
(71.92 against 68.88, per-seed standard deviation around 1), and the response falls monotonically
above three steps. On LIBERO, a single-seed check gives 82.50\% success at two steps versus 83.75\% at ten. That check has no error bars, at 100 episodes per
suite on one seed, and the LIBERO response is not monotone, with seven steps highest at 85.75, so
we treat it as indicative and rest the choice on the MetaWorld sweep.

\begin{table}[t]
\caption{Success rate against the number of flow-matching steps, SmolVLA. MetaWorld uses seeds
1000--5000 with a 500-step episode budget and 500 episodes per seed, 10 on each of the 50 tasks,
with per-seed standard deviations. LIBERO is a single seed,
100 episodes per suite over the four suites, and carries no error bars.}
\label{tab:steps}
\begin{center}
\begin{tabular}{lrr}
\toprule
\multicolumn{1}{c}{\bf FLOW-MATCHING STEPS} &\multicolumn{1}{c}{\bf METAWORLD}
&\multicolumn{1}{c}{\bf LIBERO} \\
\midrule
1            & $64.88 \pm 1.40$ & 75.25 \\
2 (ours)     & $71.92 \pm 1.29$ & 82.50 \\
3            & $72.04 \pm 0.57$ & 85.00 \\
5            & $71.60 \pm 0.93$ & 85.00 \\
7            & $70.48 \pm 0.97$ & 85.75 \\
10 (default) & $68.88 \pm 0.73$ & 83.75 \\
\bottomrule
\end{tabular}
\end{center}
\end{table}

\subsection{Control success against step budget}
\label{app:budget}

Backtracking spends steps, so we evaluate at twice the wrapper step limit and charge retreat steps
against the same budget for control and treatment alike. That raises an obvious question: how much
of the uplift could be had by simply allowing more time? Table~\ref{tab:budget} answers it by
re-scoring the control arms at intermediate budgets, which needs no additional rollouts, since an
episode that succeeded at step $n$ also succeeds under any budget of at least $n$.

Doubling the budget from the wrapper default increases success by $2.3$--$3.7$ percentage
points. Most failures at the default budget persist at the doubled budget. Backtrack's reported
uplift is measured against this longer-budget control, so it is additional to these gains from time alone.

\begin{table}[t]
\caption{Control success rate against step budget, LIBERO-Long, five seeds, 500 episodes
per seed. 520 steps is the wrapper default and 1040 is the budget used throughout the paper.
Computed offline from the per-episode step counts of the control arms, so no additional rollouts
were run; the 1040 column reproduces the control success rates in Table~\ref{tab:main}.}
\label{tab:budget}
\begin{center}
\begin{tabular}{lrrrrrrr}
\toprule
\multicolumn{1}{c}{\bf POLICY} & 520 & 624 & 728 & 832 & 936 & 1040
&\multicolumn{1}{c}{\bf $\Delta$} \\
\midrule
SmolVLA     & 57.28 & 58.64 & 59.40 & 60.20 & 60.80 & 61.00 & $+3.72$ \\
$\pi$0-FAST & 83.00 & 83.76 & 84.12 & 84.24 & 85.12 & 85.32 & $+2.32$ \\
$\pi$0.5    & 91.88 & 93.36 & 93.88 & 94.44 & 94.64 & 94.72 & $+2.84$ \\
\bottomrule
\end{tabular}
\end{center}
\end{table}

\subsection{Per-suite composition of the head training data}
\label{app:headdata}

Table~\ref{tab:headdata} in Section~\ref{sec:method} gives the size of each rollout pool.
Table~\ref{tab:headsuite} breaks those pools down by suite. All three cover all four suites, but
the mix differs substantially between policies: SmolVLA's pool is weighted toward
LIBERO-Spatial and $\pi$0.5's toward LIBERO-Long, reflecting when each set of
rollouts was collected rather than any deliberate design. The balanced sampler
(Section~\ref{sec:head}) weights every combination of suite and outcome equally, so it removes
this imbalance as well as the uneven distribution of failures across suites. Without it,
differences between the three heads would partly reflect how much data each pool has from each
suite rather than the policies themselves.

\begin{table}[t]
\caption{Rollout episodes per suite in each policy's failure-head training pool. All counts are
episodes of the frozen policy, pooled across seeds. $\pi$0-FAST's LIBERO-Long count is 47
episodes short of a round 2{,}000 because one collection run ended early.}
\label{tab:headsuite}
\begin{center}
\begin{tabular}{lrrrrr}
\toprule
\multicolumn{1}{c}{\bf POLICY} &\multicolumn{1}{c}{\bf SPATIAL} &\multicolumn{1}{c}{\bf OBJECT}
&\multicolumn{1}{c}{\bf GOAL} &\multicolumn{1}{c}{\bf LONG}
&\multicolumn{1}{c}{\bf TOTAL} \\
\midrule
SmolVLA     & 8{,}000 & 4{,}000 & 4{,}000 & 4{,}000 & 20{,}000 \\
$\pi$0-FAST & 2{,}000 & 2{,}000 & 2{,}000 & 1{,}953 &  7{,}953 \\
$\pi$0.5    & 3{,}000 & 3{,}000 & 3{,}000 & 5{,}000 & 14{,}000 \\
\bottomrule
\end{tabular}
\end{center}
\end{table}

\subsection{Feature taps evaluated}
\label{app:feature-taps}

Table~\ref{tab:taps} lists every feature tap we evaluated in closed loop, for each policy, at that
policy's headline controller configuration on LIBERO-Long. For SmolVLA and $\pi$0-FAST,
the tap comparison comes from screening runs at 10 episodes per task, the only protocol at which
every tap was run for those policies; the headline \texttt{post} arm at the full 50 episodes per
task used for the main LIBERO recovery results is listed alongside. On $\pi$0.5 the
taps were compared directly at 50 episodes per task. Each arm is paired by seed against the control
it was run with. $\pi$0-FAST has no action expert, so \texttt{post\_expert} does not exist for it,
and \texttt{pre} was never evaluated in closed loop on $\pi$0.5.

\begin{table}[t]
\caption{Feature taps evaluated per policy, at each policy's headline controller configuration on
LIBERO-Long, five seeds unless noted. Rows at 10 episodes per task are screening runs,
comparable only with the other 10-episode rows of the same policy; rows marked (headline) are the
full-protocol arms reported in Table~\ref{tab:main}. Uplift is in percentage points, paired by
seed against each arm's own control; intervals are 95\% confidence intervals over the per-seed
differences. Controls differ between runs, so compare uplifts rather than success rates across
rows.}
\label{tab:taps}
\begin{center}
\begin{tabular}{llrrrr}
\toprule
\multicolumn{1}{c}{\bf POLICY} &\multicolumn{1}{c}{\bf TAP} &\multicolumn{1}{c}{\bf EPS/TASK}
&\multicolumn{1}{c}{\bf CONTROL} &\multicolumn{1}{c}{\bf UPLIFT} &\multicolumn{1}{c}{\bf 95\% CI} \\
\midrule
SmolVLA     & \texttt{pre}         & 10 & 62.20 & $+3.00$ & $[-3.33, +9.33]$ \\
            & \texttt{post} (3 seeds) & 10 & 62.67 & $+8.67$ & $[-0.72, +18.06]$ \\
            & \texttt{post\_expert}& 10 & 60.40 & $+8.20$ & $[+0.28, +16.12]$ \\
            & \texttt{post} (headline) & 50 & 61.00 & $+7.40$ & $[+4.52, +10.28]$ \\[4pt]
$\pi$0-FAST & \texttt{pre}         & 10 & 85.20 & $+5.80$ & $[+0.08, +11.52]$ \\
            & \texttt{post}        & 10 & 85.20 & $+7.20$ & $[+4.37, +10.03]$ \\
            & \texttt{post\_expert}& \multicolumn{4}{l}{not applicable, no action expert} \\
            & \texttt{post} (headline) & 50 & 85.32 & $+7.08$ & $[+5.91, +8.25]$ \\[4pt]
$\pi$0.5    & \texttt{pre}         & \multicolumn{4}{l}{not run in closed loop} \\
            & \texttt{post} (headline) & 50 & 94.72 & $+1.60$ & $[+0.06, +3.14]$ \\
            & \texttt{post\_expert}& 50 & 94.72 & $+1.88$ & $[+0.16, +3.60]$ \\
\bottomrule
\end{tabular}
\end{center}
\end{table}

\subsection{Executable reversal and privileged restoration}
\label{app:restoration}

Backtrack returns the arm by physically replaying its recorded motion in reverse. MetaWorld also
allows a privileged alternative: restoring the arm directly to its earlier state without executing
the return trajectory. At the same operating point, instantaneous restoration achieves
$+5.72$\,pp compared with $+2.80$\,pp for executable reversal (Table~\ref{tab:teleport}); the
seed-paired difference is
$+2.92$\,pp with a 95\% interval of $[+1.44,+4.40]$.

This comparison does not isolate the source of the gap. Instantaneous restoration both returns the
arm more precisely and avoids the time spent executing the retreat, during which the scene may
continue to evolve. It therefore represents a privileged simulator intervention rather than a
variant of Backtrack that could be executed directly on a physical system.

\begin{table}[t]
\caption{Executable reversal against privileged instantaneous restoration, SmolVLA on MetaWorld,
seeds 6000--10000, a 500-step episode budget, and 500 episodes per seed. Both recovery mechanisms use the same operating point
($t{=}0.8$, 100-step retreat, 100-step cooldown) and are paired by seed against a shared control.
All three arms come from the same evaluation sweep; uplifts are relative to its control.}
\label{tab:teleport}
\begin{center}
\begin{tabular}{lrrr}
\toprule
\multicolumn{1}{c}{\bf RECOVERY} &\multicolumn{1}{c}{\bf SR}
&\multicolumn{1}{c}{\bf UPLIFT} &\multicolumn{1}{c}{\bf 95\% CI} \\
\midrule
control                       & 70.56 &         &                  \\
executable reversal           & 73.36 & $+2.80$ & $[+1.87, +3.73]$ \\
instantaneous restoration     & 76.28 & $+5.72$ & $[+3.37, +8.07]$ \\
\bottomrule
\end{tabular}
\end{center}
\end{table}

\subsection{Retreat-cap ablation}
\label{app:cap}

\emph{Maximum retreats per episode} limits how many times the controller may initiate a
retreat during an episode. Once an episode reaches this cap, the trigger is disabled for the remainder of that
episode and the policy runs unassisted to the step limit. The cap does not change the length of an
individual retreat, only how many times the controller may fire.

We vary the cap over 1, 2, 3, 4 and 5 on $\pi$0.5 with the \texttt{post} head, on
LIBERO-Long, holding the rest of the headline configuration fixed: a fixed 50-step retreat,
a 50-step cooldown, and a trigger threshold of $t=0.99$. Each arm is three seeds (1000, 2000, 3000)
at 50 episodes per task over the suite's ten tasks, so 500 episodes per seed and 1{,}500 per cap,
evaluated at twice the wrapper step limit (1040 steps). Retreat steps are charged against that
budget, as everywhere else in the paper.

\begin{table}[t]
\caption{Retreat cap on $\pi$0.5, LIBERO-Long, \texttt{post} head, fixed 50-step retreat,
50-step cooldown, $t=0.99$. Per-seed success rates at 50 episodes per task, and the seed-paired
uplift against the shared control. Intervals are 95\% confidence intervals over the three per-seed
differences, using $t_{0.975,2}=4.303$; they are wide because $n=3$, and the comparison of interest
is between caps rather than against zero. Control is the main evaluation's control restricted to
seeds 1000--3000, giving 94.07\% success and a $+2.20$\,pp uplift for the headline configuration,
compared with 94.72\% and $+1.60$\,pp over all five seeds (Table~\ref{tab:main}). The independently
evaluated cap-3 arm matches the headline arm's per-seed success rates on these three seeds.}
\label{tab:cap}
\begin{center}
\begin{tabular}{lrrrrrrr}
\toprule
&\multicolumn{3}{c}{\bf SUCCESS RATE BY SEED} \\
\cmidrule(lr){2-4}
\multicolumn{1}{c}{\bf CAP} & 1000 & 2000 & 3000 &\multicolumn{1}{c}{\bf MEAN}
&\multicolumn{1}{c}{\bf SD} &\multicolumn{1}{c}{\bf UPLIFT} &\multicolumn{1}{c}{\bf 95\% CI} \\
\midrule
control & 93.20 & 95.00 & 94.00 & 94.07 & 0.90 &        &  \\[4pt]
1       & 96.80 & 95.60 & 95.40 & 95.93 & 0.76 & $+1.87$ & $[-1.99, +5.73]$ \\
2       & 96.60 & 95.20 & 95.40 & 95.73 & 0.76 & $+1.67$ & $[-2.35, +5.68]$ \\
3       & 96.20 & 97.00 & 95.60 & 96.27 & 0.70 & $+2.20$ & $[+0.41, +3.99]$ \\
4       & 96.80 & 96.00 & 95.80 & 96.20 & 0.53 & $+2.13$ & $[-1.17, +5.44]$ \\
5       & 95.40 & 96.00 & 95.60 & 95.67 & 0.31 & $+1.60$ & $[+0.11, +3.09]$ \\
\bottomrule
\end{tabular}
\end{center}
\end{table}

Table~\ref{tab:cap} gives the result. The five caps span 0.60 percentage points, from 95.67 at a
cap of 5 to 96.27 at a cap of 3, against a per-seed standard deviation of 0.31 to 0.76 within each
arm. Comparing caps directly and pairing by seed, a cap of 1 against a cap of 3 is $-0.33$\,pp with
an interval of $[-2.83, +2.17]$, and a cap of 3 against a cap of 5 is $+0.60$\,pp with an interval
of $[-0.71, +1.91]$. Neither comparison resolves a difference. With only three seeds,
these intervals still allow differences of several percentage points; they do not establish
that one retreat is sufficient or that the caps are equivalent.

\subsection{MetaWorld controller sweep}
\label{app:metaworld-sweep}

Table~\ref{tab:metaworld-sweep} reports all twelve controller configurations evaluated in
Section~\ref{sec:metaworld}. Each configuration is compared with the control from this sweep.
The flow-step ablation (Table~\ref{tab:steps}) and restoration comparison
(Table~\ref{tab:teleport}) are separate evaluations with 500-step budgets; this sweep uses
1000 steps for both control and treatment.

\begin{table}[t]
\caption{Controller sweep on MetaWorld, SmolVLA, \texttt{pre} head. All arms use seeds
1000--5000, a 1000-step episode budget including retreat steps, and 500 episodes per seed
(10 per task across all 50 tasks). Control success is 72.32\%. SR is success rate (\%);
uplifts and intervals are in percentage points relative to this sweep's control.
Intervals are 95\% confidence intervals over the five per-seed differences. Here $t$ is the trigger threshold, $a$ the adaptive
restart threshold, $s$ the fixed retreat length or adaptive minimum--maximum, and $cd$ the
cooldown; lengths and cooldowns are in steps.}
\label{tab:metaworld-sweep}
\begin{center}
\begin{tabular}{rrrrrrr}
\toprule
$t$ & $a$ & $s$ & $cd$ & \textbf{SR} & \textbf{UPLIFT} & \textbf{95\% CI} \\
\midrule
\multicolumn{4}{l}{Control} & 72.32 & & \\[4pt]
\multicolumn{7}{l}{Fixed retreat} \\[2pt]
0.8 & -- & 100 & 50  & 74.76 & $+2.44$ & $[-0.65, +5.53]$ \\
0.8 & -- & 100 & 100 & 76.28 & $+3.96$ & $[+1.62, +6.30]$ \\
0.9 & -- & 100 & 50  & 75.44 & $+3.12$ & $[+1.14, +5.10]$ \\
0.9 & -- & 100 & 100 & 75.96 & $+3.64$ & $[+1.57, +5.71]$ \\
0.9 & -- & 150 & 100 & 77.52 & $+5.20$ & $[+2.28, +8.12]$ \\
0.9 & -- & 200 & 100 & 77.20 & $+4.88$ & $[+1.61, +8.15]$ \\[4pt]
\multicolumn{7}{l}{Adaptive retreat} \\[2pt]
0.8 & 0.2 & 50--150 & 50  & 75.32 & $+3.00$ & $[+0.37, +5.63]$ \\
0.9 & 0.1 & 50--150 & 50  & 75.28 & $+2.96$ & $[+1.34, +4.58]$ \\
0.9 & 0.2 & 50--100 & 50  & 75.92 & $+3.60$ & $[+0.57, +6.63]$ \\
0.9 & 0.2 & 50--150 & 50  & 75.96 & $+3.64$ & $[+1.09, +6.19]$ \\
0.9 & 0.2 & 50--200 & 50  & 76.44 & $+4.12$ & $[+1.38, +6.86]$ \\
0.9 & 0.3 & 50--100 & 100 & 76.36 & $+4.04$ & $[+1.74, +6.34]$ \\
\bottomrule
\end{tabular}
\end{center}
\end{table}

\end{document}